\documentclass[11pt,anonymous=false,a4paper]{article}
\usepackage{paclic34}
\usepackage{times}
\usepackage{latexsym}
\usepackage{amsmath}
\usepackage{multirow}
\usepackage{url}

\title{UIT-ViCoV19QA: A Dataset for COVID-19 Community-based \\Question Answering on Vietnamese Language}
\usepackage[utf8]{vietnam}
\usepackage[utf8]{inputenc}
\usepackage{caption}
\usepackage{booktabs}
\usepackage{multirow}
\usepackage{float}
\usepackage[table,xcdraw]{xcolor}
\usepackage[vietnamese]{babel}
\usepackage{graphicx}
\usepackage{svg}
\usepackage{hyperref}
\usepackage{xcolor,soulutf8}

\definecolor{OliveGreen}{HTML}{FFB3B3}
\sethlcolor{OliveGreen}

\author{
    Triet Minh Thai\\
    University of Information Technology\\VNU-HCM, Vietnam\\ {\tt 19522397@gm.uit.edu.vn}
    \And
    Ngan Ha-Thao Chu\\
    University of Information Technology\\VNU-HCM, Vietnam\\ {\tt 19521882@gm.uit.edu.vn}
    \AND
    Anh Tuan Vo\\
    University of Information Technology\\VNU-HCM, Vietnam\\ {\tt 19521226@gm.uit.edu.vn}
    \And
    Son T. Luu\\
    University of Information Technology\\VNU-HCM, Vietnam\\ {\tt sonlt@uit.edu.vn}
    }

\date{}

\begin{document}
\maketitle
\begin{abstract}
For the last two years, from 2020 to 2021, COVID-19 has broken disease prevention measures in many countries, including Vietnam, and negatively impacted various aspects of human life and the social community. Besides, the misleading information in the community and fake news about the pandemic are also serious situations. Therefore, we present the first Vietnamese community-based question answering dataset for developing question answering systems for COVID-19 called UIT-ViCoV19QA. The dataset comprises 4,500 question-answer pairs collected from trusted medical sources, with at least one answer and at most four unique paraphrased answers per question. Along with the dataset, we set up various deep learning models as baseline to assess the quality of our dataset and initiate the benchmark results for further research through commonly used metrics such as BLEU, METEOR, and ROUGE-L. We also illustrate the positive effects of having multiple paraphrased answers experimented on these models, especially on Transformer - a dominant architecture in the field of study.

\end{abstract}

\section{Introduction}

Community-based Question answering (CQA) is a task of question answering based on the wisdom of the crowd \cite{zhang2021graph}. In CQA, information seekers post their questions on a public website or forum, and other users answer them. This kind of question-answering behavior is popular in people's daily basis. For example, Quora\footnote{\url{https://www.quora.com/}} and Reddit\footnote{\url{https://www.reddit.com/}} are several large forums for question-answering. The CQA enables people to ask and answer questions easily \cite{qiu2015convolutional}. Additionally, the development of question and answering systems means computers can now understand and answer the questions of users. 

In Vietnam, the year from 2020 to 2021 witnessed the COVID-19 pandemic. Information about the COVID-19 spreading situation, medical care, self-quarantine, vaccination policies, and regulations by the government to prevent the spread of COVID-19 are essential to citizens. People frequently ask questions about the COVID-19 situation, what to do when contacting COVID-19 patients, the vaccination policies, and more. This is our motivation to construct a dataset 
to help build a question answering system based on CQA about COVID-19 in Vietnamese. Apart from the dataset, we also propose various baseline models to evaluate our dataset's quality. This paper has three main contributions summarized as follows:

\begin{enumerate}
  \item We introduce UIT-ViCoV19QA, the first community-based question answering collection about the COVID-19 pandemic for Vietnamese constructed from trusted sources. The dataset comprises 4,500 question-answer pairs and is extended to have up to four unique paraphrased answers per question through an efficient paraphrasing process.
  
  \item We assess the dataset's quality and establish a future research benchmark through experiments with various Sequence-to-Sequence baselines to automatically generate answer for a given question about COVID-19 in Vietnamese.
  
  \item We perform error analysis and illustrate that models trained on multiple paraphrased answers tend to have better generalization than those trained using only one answer. This reflects the advantage of having multiple paraphrased answers in single-turn conversational question answering.
\end{enumerate}

The rest of this paper is structured as follows. In the following section, we review the related works. In Section 3, we describe the process of building UIT-ViCoV19QA dataset in detail, including data collection, data pre-processing, paraphrasing process, and highlight its overall statistics. Section 4 is devoted to methodologies and experiment configurations. The results and benchmarks, as well as error analysis are described in Section 5. Finally, our conclusion and future works are presented in Section 6.
\section{Related works}
Question answering systems consist of the single-turn and multi-turn QA. According to \cite{del2021question}, the single-turn QA takes the questions as input and returns the output without context. Single-turn QA includes Text-based QA (SQuAD \cite{rajpurkar-etal-2016-squad}, RACE \cite{lai-etal-2017-race}),
Visual-QA (VQA \cite{goyal2017making}), Community-based QA (ANTIQUE \cite{hashemi2020antique}), and Knowledge-based QA (MetaQA \cite{zhang2018variational}) In contrast, the multi-turn QA take the questions as input belong with contexts such as conversation history, which is called the Conversational QA (CoQA \cite{reddy-etal-2019-coqa}, QuAC \cite{choi-etal-2018-quac}). Additionally, to verbalize the response from the question, the ParaQA dataset \cite{kacupaj2021paraqa} used the paraphrasing techniques. Each question in the ParaQA dataset has at least two more answers. In particular, some questions contain eight responses.

On the other hand, many efforts to create the question answering corpora in Vietnamese such as UIT-ViQuAD \cite{nguyen2020vietnamese}, UIT-ViNewsQA \cite{van2020new}, ViMMRC \cite{van2020enhancing} for text-based question answering, ViVQA \cite{tran2021vivqa} for visual-based QA, and ViCoQA \cite{luu2021conversational} for conversational QA. Our works contributes to the corpora for Vietnamese as Community-based QA dataset. With the idea from the ParaQA dataset \cite{kacupaj2021paraqa}, we manually create up to three responses from initial answer for each question
in our dataset to make the dataset verbalized. 
\section{Dataset} 
\begin{figure*}[ht]
\centering
\includegraphics[width=.9\textwidth]{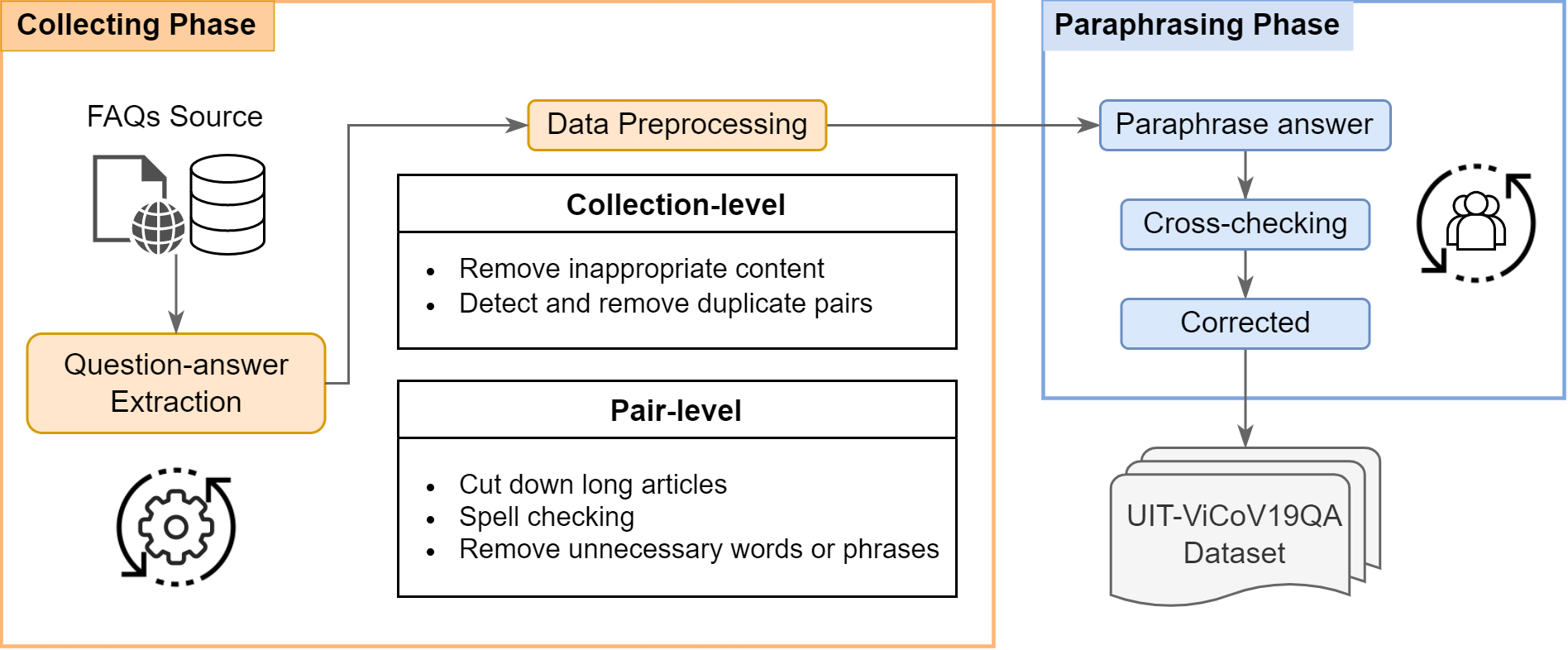}
\caption{An overview of UIT-ViCoV19QA constructing workflow}
\label{fig_workflow}
\end{figure*}

In order to publish a high-quality dataset for the research community and be able to experiment with the baseline models, we have investigated building a dataset for Vietnamese, a low-resource language in the field of study. This section describes the construction of  UIT-ViCoV19QA in detail and presents some statistics from the dataset. The overview of the dataset construction workflow is illustrated in Figure \ref{fig_workflow}.

\subsection{Data Collection}
 Question-answer pairs used for the constructing of UIT-ViCoV19QA dataset are extracted from FAQ documents that are publicly available on websites of respected health care organizations in Vietnam and overseas, including The Centers for Disease Control and Prevention (CDC), United Nations Children's Fund (UNICEF), The Ministry of Health of Vietnam, Vietnam Government Portal and other trusted medical institutions. Each web page provides different topics about the COVID-19 pandemic compiled in Vietnamese and often includes similar information extracted during the crawl. The following topics are covered in our dataset: origin, outbreak, and name of the disease; spread; symptoms; prevention, treatment guidelines, and nutrition; treatment models; variants of COVID-19; vaccines and vaccination; moving between areas, entry, and travel; isolation, quarantine, lockdown and social distancing; policies and sanctions; financial support; post-COVID-19; COVID-19 in children. Disease statistics in Vietnam and worldwide are not included in the dataset.

Once appropriate and trustful sources are identified, numerous handcrafted patterns are developed for each website to automatically extract question-answer pairs based on the document structure, typically in HTML and PDF format. Throughout the collecting phase, the length of both question and answers are maintained to retain information. Therefore, the collected content could be a sentence, a paragraph, or a passage determined by the author or compiler. This poses a tremendous challenge for Sequence-to-Sequence models to produce good results, which makes it an ideal material for further investigations.

\subsection{Data Pre-processing}
A two-stage pre-processing is performed on the collected question-answer pairs. First, duplicate pairs are detected using Cosine similarity at the collection-level and are considered to be taken out of the collection. During the crawl, we also attempt to identify and remove pairs that have inappropriate content, such as articles that are neither related to COVID-19 nor the Vietnamese community or are too complicated to comprehend. 

Second, at the pair-level, we check the spelling and correct mistakes. Undesirable artifacts, erroneous character and HTML tags will also be removed in this stage. In case the collected articles are too long or contains more than three paragraphs, extracted answers will be manually reduced in length by removing unnecessary information, such as source name, footnote, greetings, preliminary, over-specified explanations, and farewell. After this step, 4,500 Vietnamese question-answer pairs about  COVID-19 are achieved from trusted FAQ sources. Some examples of the collected question-answer pairs are shown in Table \ref{tbl_examples}.

\begin{table*}[ht]
\renewcommand{\arraystretch}{1.2}
\begin{center}
\begin{tabular}{p{0.95\linewidth}}

\toprule
\textbf{Question:} Xử lý triệu chứng ho khi chăm sóc F0 tại nhà thế nào? [\textbf{English: }How to handle cough symptoms when taking care of F0 patient at home?]\\
\textbf{Answer:} Dùng thuốc giảm ho theo đơn của bác sĩ. Có thể dùng thêm các vitamin theo đơn thuốc của bác sĩ. [\textbf{English: }Take cough suppressants as prescribed by your doctor. Additional vitamins can be taken according to the doctor's prescription.]\\ 
\midrule
\textbf{Question:} Tôi đang điều trị viêm tiết niệu và viêm dạ dày hội chứng ruột kích thích thì có tiêm vaccine Covid-19 được hay không? [\textbf{English: }I am being treated for UTIs and gastritis with irritable bowel syndrome, can I get the Covid-19 vaccine?]\\
\textbf{Answer:} Với bệnh cấp tính mà anh/chị đang mắc phải cần được điều trị ổn định trước, sức khỏe tốt, bình thường thì có thể tiêm vaccine Covid-19. [\textbf{English: }If you have an acute illness that are stably treated, if you are in good health, you can receive the Covid-19 vaccine.]\\
\midrule
\textbf{Question:} Hỗ trợ hô hấp cho trẻ em nhiễm COVID-19 ở thế nặng như thế nào? [\textbf{English: }How to provide respiratory support to children with COVID-19 in severe condition?]\\
\textbf{Answer:} Thở mask có túi Hoặc: NCPAP, HPNO, NIPPV [\textbf{English: }Apply breathing mask with bag or: NCPAP, HPNO, NIPPV]\\\bottomrule
\end{tabular}
\end{center}
\caption{Examples of question-answer pairs from UIT-ViCoV19QA}
\label{tbl_examples} 
\end{table*}

\subsection{Paraphrase Generation Process}
Inspired by the concept of \newcite{kacupaj2021paraqa}, we have investigated extending our dataset using the following paraphrasing methods on Vietnamese samples:

\begin{itemize}
    \item Rearrange words, phrases, or sentences in the initial answer to create new responses without changing their meaning.
    \item Reduce or diversify the content of the initial answer.
    \item Paraphrase the initial answer using synonyms and similar structures.
\end{itemize}

These methods are manually applied on the collection to create up to three individual paraphrased responses consecutively for each question-answer pair. Newly created answers will be annotated in order to indicate the minimum number of answers per question.

After creating multiple paraphrased versions of the initial answer, we perform a cross-checking process to correct the spelling mistakes and grammatical errors as well as modify and rephrase digressive answers. By the end of this phase, the UIT-ViCoV19QA dataset is entirely constructed with 4,500 question-answer pairs containing at least one answer and at most four unique paraphrased answers per question.

\subsection{Statistics}
The statistics of the training, development, and test sets are described in Table \ref{tbl_stat_dataset}. The UIT-ViCoV19QA dataset consists of 4,500 question-answer pairs in total. In the table, the average length, as well as the vocabulary size \footnote{We use underthesea package: \url{https://github.com/undertheseanlp/underthesea} for word segmentation.} of questions and answers, are also presented.

\begin{figure}[ht]
\centering
\hspace*{-0.4cm}\includegraphics[width=0.5
\textwidth]{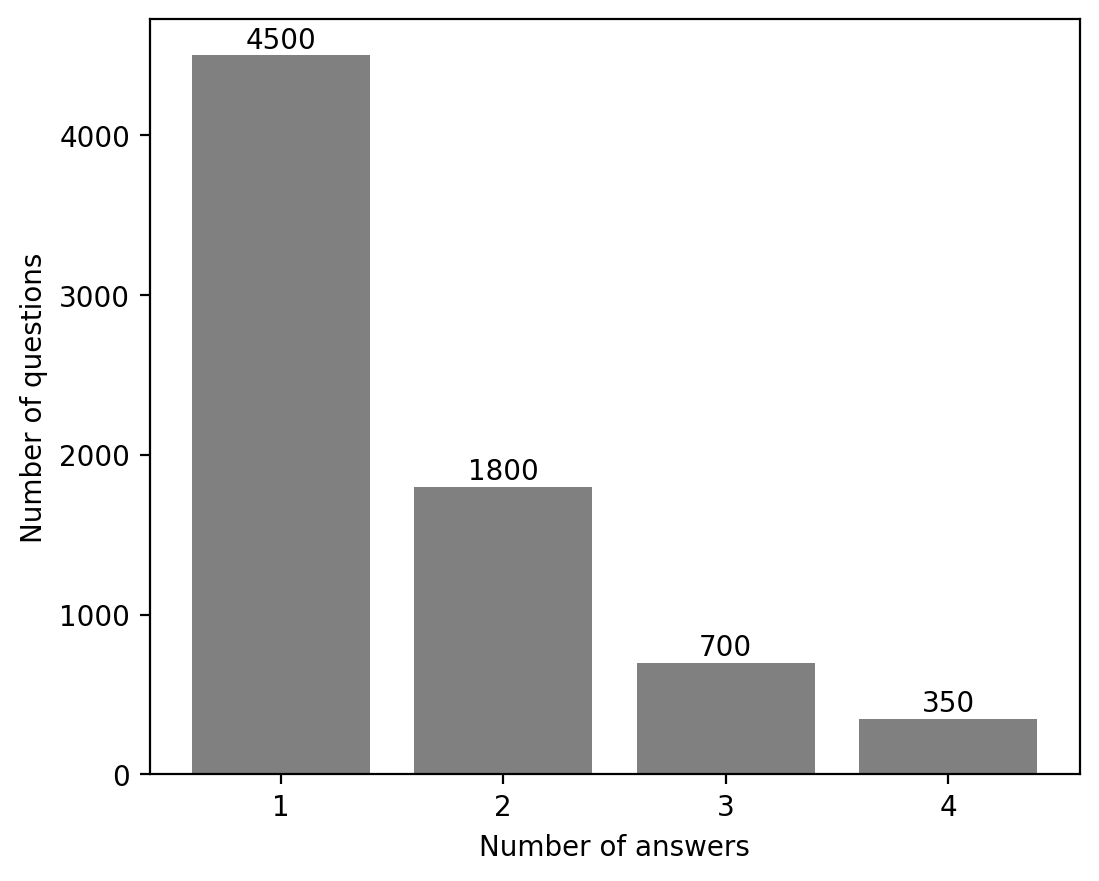}
\setlength{\belowcaptionskip}{-10pt}
\caption{Distribution of total number of answers per question in UIT-ViCoV19QA}
\label{fig_distribute_response}
\end{figure}

Figure \ref{fig_distribute_response} illustrates the distribution of 4,500 questions of UIT-ViCoV19QA based on number of answers per question. The figure shows that the dataset contains 1800 questions that have at least two answers, 700 questions have at least three answers and half of them have a maximum of four paraphrased answers.

\begin{table*}[ht]
\centering
\begin{tabular}{@{}llrrrr@{}}
\toprule
\multicolumn{2}{l}{}                                                               & \multicolumn{1}{c}{\textbf{Train}}        & \multicolumn{1}{c}{\textbf{Dev.}}         & \multicolumn{1}{c}{\textbf{Test}}        & \multicolumn{1}{c}{\textbf{All}}          \\ \midrule
{\color[HTML]{000000} }                           & Number of question-answer pairs & 3500                             & 500                             & 500                             & 4500                             \\
{\color[HTML]{000000} }                           & Average question length        & \cellcolor[HTML]{FFFFFF}31.44   & \cellcolor[HTML]{FFFFFF}33.66   & \cellcolor[HTML]{FFFFFF}32.32  & \cellcolor[HTML]{FFFFFF}31.79  \\
{\color[HTML]{000000} }                           & Average answer length          & \cellcolor[HTML]{FFFFFF}120.53 & \cellcolor[HTML]{FFFFFF}116.04 & \cellcolor[HTML]{FFFFFF}118.11 & \cellcolor[HTML]{FFFFFF}119.76 \\
{\color[HTML]{000000} }                           & Question vocabulary size       & \cellcolor[HTML]{FFFFFF}4396     & \cellcolor[HTML]{FFFFFF}1869    & \cellcolor[HTML]{FFFFFF}1770    & \cellcolor[HTML]{FFFFFF}4924     \\
\multirow{-5}{*}{{\color[HTML]{000000} \textbf{Answer 1}}} & Answer vocabulary size         & \cellcolor[HTML]{FFFFFF}8537     & \cellcolor[HTML]{FFFFFF}3689    & \cellcolor[HTML]{FFFFFF}3367    & \cellcolor[HTML]{FFFFFF}9411     \\\midrule
{\color[HTML]{000000} }                           & Number of question-answer pairs & \cellcolor[HTML]{FFFFFF}1390     & 209                             & 201                             & 1800                             \\
{\color[HTML]{000000} }                           & Average question length        & \cellcolor[HTML]{FFFFFF}35.56  & \cellcolor[HTML]{FFFFFF}39.22 & \cellcolor[HTML]{FFFFFF}39.72 & \cellcolor[HTML]{FFFFFF}36.45  \\
{\color[HTML]{000000} }                           & Average answer length          & \cellcolor[HTML]{FFFFFF}40.54   & \cellcolor[HTML]{FFFFFF}39.25 & \cellcolor[HTML]{FFFFFF}42.73 & \cellcolor[HTML]{FFFFFF}40.64  \\
{\color[HTML]{000000} }                           & Question vocabulary size       & \cellcolor[HTML]{FFFFFF}2883     & \cellcolor[HTML]{FFFFFF}1269    & \cellcolor[HTML]{FFFFFF}1207    & \cellcolor[HTML]{FFFFFF}3305     \\
\multirow{-5}{*}{{\color[HTML]{000000} \textbf{Answer 2}}} & Answer vocabulary size         & \cellcolor[HTML]{FFFFFF}2632     & \cellcolor[HTML]{FFFFFF}1098    & \cellcolor[HTML]{FFFFFF}1129    & \cellcolor[HTML]{FFFFFF}2949     \\\midrule
{\color[HTML]{000000} }                           & Number of question-answer pairs & 542                              & 79                              & 79                              & 700                              \\
{\color[HTML]{000000} }                           & Average question length        & \cellcolor[HTML]{FFFFFF}34.77  & \cellcolor[HTML]{FFFFFF}36.70 & \cellcolor[HTML]{FFFFFF}39.28 & \cellcolor[HTML]{FFFFFF}35.49  \\
{\color[HTML]{000000} }                           & Average answer length          & \cellcolor[HTML]{FFFFFF}28.68  & \cellcolor[HTML]{FFFFFF}26.43 & \cellcolor[HTML]{FFFFFF}30.89 & \cellcolor[HTML]{FFFFFF}28.67  \\
{\color[HTML]{000000} }                           & Question vocabulary size       & \cellcolor[HTML]{FFFFFF}1836     & \cellcolor[HTML]{FFFFFF}717     & \cellcolor[HTML]{FFFFFF}693     & \cellcolor[HTML]{FFFFFF}2111     \\
\multirow{-5}{*}{{\color[HTML]{000000} \textbf{Answer 3}}} & Answer vocabulary size         & \cellcolor[HTML]{FFFFFF}1554     & \cellcolor[HTML]{FFFFFF}503     & \cellcolor[HTML]{FFFFFF}585     & \cellcolor[HTML]{FFFFFF}1753     \\\midrule
{\color[HTML]{000000} }                           & Number of question-answer pairs & 272                              & 39                              & 39                              & 350                              \\
{\color[HTML]{000000} }                           & Average question length        & \cellcolor[HTML]{FFFFFF}36.57  & \cellcolor[HTML]{FFFFFF}37.59 & \cellcolor[HTML]{FFFFFF}42.15 & \cellcolor[HTML]{FFFFFF}37.10  \\
{\color[HTML]{000000} }                           & Average answer length          & \cellcolor[HTML]{FFFFFF}29.75  & \cellcolor[HTML]{FFFFFF}29.03 & \cellcolor[HTML]{FFFFFF}35.72 & \cellcolor[HTML]{FFFFFF}30.25  \\
{\color[HTML]{000000} }                           & Question vocabulary size       & \cellcolor[HTML]{FFFFFF}1315     & \cellcolor[HTML]{FFFFFF}470     & \cellcolor[HTML]{FFFFFF}460     & \cellcolor[HTML]{FFFFFF}1519     \\
\multirow{-5}{*}{{\color[HTML]{000000} \textbf{Answer 4}}} & Answer vocabulary size         & \cellcolor[HTML]{FFFFFF}924      & \cellcolor[HTML]{FFFFFF}353     & \cellcolor[HTML]{FFFFFF}374     & \cellcolor[HTML]{FFFFFF}1075     \\ \midrule
\end{tabular}
\caption{Overall statistics of the UIT-ViCoV19QA dataset.}
\label{tbl_stat_dataset}
\end{table*}

\section{Methodologies}

\subsection{Baseline Models}
This experimental section set up various deep learning models with Encoder-Decoder architecture to evaluate the UIT-ViCoV19QA dataset. These models have achieved significant results on many sequence-to-sequence learning tasks that involve long sequences, such as machine translation, text summarization, and question answering.

\begin{itemize}

    \item \textbf{Attention-based Recurrent Neural Network}: Recurrent neural network (RNN) models used in the experiments are implemented with two attention mechanism - Bahdanau Attention \cite{bahdanau2016neural} and Luong Attention \cite{luong-etal-2015-effective}. To ensure that the results are comparable, these models are set up with similar hyperparameters in the encoder and decoder as follows: an embedding layer with dimension 512, two hidden layers of 512 gated recurrent unit (GRU) cells, and a drop-out rate of 0.5. Bidirectional gated recurrent unit (Bi-GRU) is applied in the encoder of both models to help them understand the context better. For simplicity, RNN models using Bahdanau attention and Luong attention are annotated as RNN-1 and RNN-2, respectively.
    
    \item \textbf{Convolutional Network} \cite{10.5555/3305381.3305510}:
    Different from RNN, convolutional neural network uses many convolutional layers typically applied in image processing. Each layer uses filters to learn to extract different features from the text. In our experiments, the hyperparameters of models are set as follows: embedding layer with dimension 512, three convolutional layers with hidden size 512 use 1024 filters with kernel size 3 x 3, and drop-out probability 0.5.
     
    \item \textbf{Transformer} \cite{NIPS2017_3f5ee243}: As a dominant architecture in natural language processing (NLP), the model and its variants, such as BERT and pre-trained versions of BERT, have been commonly used to achieve state-of-the-art results for many tasks in the field. The model is set up with these settings: embedding layer with dimension 512, two layers with 8 self-attention heads, positional embedding layer with max length 500, position-wise feed-forward layer with dimension 2048 and  drop-out rate 0.5.
    \end{itemize}

\subsection{Evaluation Metrics}
Three standard metrics utilized for evaluating baseline models. These metrics are commonly used in machine translation and text summarization tasks to compare the generated text with human performance.
\begin{itemize}
    \item \textbf{BLEU} \cite{papineni2002bleu}: BLEU is an n-gram based evaluation metric, widely used for Machine Translation (MT) evaluation to claim a high correlation with human judgments of quality. It aims to count the n-gram overlaps in the reference by taking the maximum count of each n-gram and clipping the count of the n-grams in the candidate text to the maximum count in the reference. In our experiments, BLEU score is calculated using unigram (BLEU-1) and 4-gram (BLEU-4) with uniform weight $w_n=0.25$.
    
    \item \textbf{METEOR} \cite{lavie-agarwal-2007-meteor}: The metric is based on the harmonic mean of unigram precision and recall, with recall weighted higher than precision. Moreover, it has several features not found in other metrics, such as stemming and synonymy matching, along with the standard exact word matching. Unfortunately, those features are not yet supported for the Vietnamese language. We set up standard METEOR only using exact matching for evaluation with $\alpha = 0.9$, $\beta = 3.0$ and $\gamma = 0.5$.
    \item \textbf{ROUGE-L} \cite{lin-2004-rouge}: This metric measures the longest common subsequence (LCS) between our model output and reference. The idea here is that a longer shared sequence would indicate more similarity between the two sequences. An advantage of using LCS is that it does not require consecutive matches but in-sequence matches that reflect sentence-level word order. For each generated answer, we choose the best score achieved from comparing it with all existing reference answers.
\end{itemize}


\subsection{Experimental Configuration}

\begin{table*}[ht]
\setlength{\tabcolsep}{4pt}
\small
\centering
\begin{tabular}{l|l|ccccc}
\toprule
\multicolumn{2}{c}{{\color[HTML]{000000} \textbf{Model}}} &
  {\color[HTML]{000000} \textbf{\# answers}} &
  {\color[HTML]{000000} \textbf{BLEU-1 (\%)}} &
  {\color[HTML]{000000} \textbf{BLEU-4 (\%)}} &
  {\color[HTML]{000000} \textbf{METEOR (\%)}} &
  {\color[HTML]{000000} \textbf{ROUGE-L (\%)}} \\\midrule
\multicolumn{2}{l}{{\color[HTML]{000000} }} &
  {\color[HTML]{000000} 1} &
  {\color[HTML]{000000} 21.79} &
  {\color[HTML]{000000} 10.29} &
  {\color[HTML]{000000} 25.34} &
  {\color[HTML]{000000} 32.36} \\
\multicolumn{2}{l}{{\color[HTML]{000000} }} &
  {\color[HTML]{000000} 2} &
  {\color[HTML]{000000} \textbf{\underline{26.62}}} &
  {\color[HTML]{000000} \textbf{12.86}} &
  {\color[HTML]{000000} 25.15} &
  {\color[HTML]{000000} 33.66} \\
\multicolumn{2}{l}{{\color[HTML]{000000} }} &
  {\color[HTML]{000000} 3} &
  {\color[HTML]{000000} 26.09} &
  {\color[HTML]{000000} 12.63} &
  {\color[HTML]{000000} \textbf{\underline{25.98}}} &
  {\color[HTML]{000000} \textbf{33.68}} \\
\multicolumn{2}{l}{\multirow{-4}{*}{{\color[HTML]{000000} RNN-1 \cite{bahdanau2016neural}}}} &
  {\color[HTML]{000000} 4} &
  {\color[HTML]{000000} 24.56} &
  {\color[HTML]{000000} 12.27} &
  {\color[HTML]{000000} 23.91} &
  {\color[HTML]{000000} 32.57} \\\midrule
\multicolumn{2}{l}{{\color[HTML]{000000} }} &
  {\color[HTML]{000000} 1} &
  {\color[HTML]{000000} 21.04} &
  {\color[HTML]{000000} 10.94} &
  {\color[HTML]{000000} \textbf{24.72}} &
  {\color[HTML]{000000} \textbf{\underline{33.95}}} \\
\multicolumn{2}{l}{{\color[HTML]{000000} }} &
  {\color[HTML]{000000} 2} &
  {\color[HTML]{000000} 24.07} &
  {\color[HTML]{000000} 11.30} &
  {\color[HTML]{000000} 23.81} &
  {\color[HTML]{000000} 31.92} \\
\multicolumn{2}{l}{{\color[HTML]{000000} }} &
  {\color[HTML]{000000} 3} &
  {\color[HTML]{000000} 23.8} &
  {\color[HTML]{000000} 10.95} &
  {\color[HTML]{000000} 23.65} &
  {\color[HTML]{000000} 31.65} \\
\multicolumn{2}{l}{\multirow{-4}{*}{{\color[HTML]{000000} RNN-2 \cite{luong-etal-2015-effective}}}} &
  {\color[HTML]{000000} 4} &
  {\color[HTML]{000000} \textbf{24.38}} &
  {\color[HTML]{000000} \textbf{12.30}} &
  {\color[HTML]{000000} 24.29} &
  {\color[HTML]{000000} 32.09} \\\midrule

\multicolumn{2}{l}{{\color[HTML]{000000} }} &
  {\color[HTML]{000000} 1} &
  {\color[HTML]{000000} 19.17} &
  {\color[HTML]{000000} 9.51} &
  {\color[HTML]{000000} \textbf{22.76}} &
  {\color[HTML]{000000} \textbf{32.29}} \\
\multicolumn{2}{l}{{\color[HTML]{000000} }} &
  {\color[HTML]{000000} 2} &
  {\color[HTML]{000000} 21.63} &
  {\color[HTML]{000000} 10.81} &
  {\color[HTML]{000000} 20.66} &
  {\color[HTML]{000000} 30.71} \\
\multicolumn{2}{l}{{\color[HTML]{000000} }} &
  {\color[HTML]{000000} 3} &
  {\color[HTML]{000000} 22.08} &
  {\color[HTML]{000000} 10.68} &
  {\color[HTML]{000000} 20.96} &
  {\color[HTML]{000000} 30.84} \\
\multicolumn{2}{l}{\multirow{-4}{*}{{\color[HTML]{000000} Convolutional \cite{10.5555/3305381.3305510}}}} &
  {\color[HTML]{000000} 4} &
  {\color[HTML]{000000} \textbf{23.26}} &
  {\color[HTML]{000000} \textbf{10.93}} &
  {\color[HTML]{000000} 21.84} &
  {\color[HTML]{000000} 31.15} \\\midrule
 
\multicolumn{2}{l}{{\color[HTML]{000000} }} &
  {\color[HTML]{000000} 1} &
  {\color[HTML]{000000} 21.84} &
  {\color[HTML]{000000} 10.82} &
  {\color[HTML]{000000} 23.37} &
  {\color[HTML]{000000} 31.67} \\
\multicolumn{2}{l}{{\color[HTML]{000000} }} &
  {\color[HTML]{000000} 2} &
  {\color[HTML]{000000} 24.81} &
  {\color[HTML]{000000} 13.22} &
  {\color[HTML]{000000} \textbf{24.60}} &
  {\color[HTML]{000000} \textbf{32.25}} \\
\multicolumn{2}{l}{{\color[HTML]{000000} }} &
  {\color[HTML]{000000} 3} &
  {\color[HTML]{000000} 24.90} &
  {\color[HTML]{000000} 13.64} &
  {\color[HTML]{000000} 21.72} &
  {\color[HTML]{000000} 32.20} \\
\multicolumn{2}{l}{\multirow{-4}{*}{{\color[HTML]{000000} Transformer \cite{NIPS2017_3f5ee243}}}} &
  {\color[HTML]{000000} 4} &
  {\color[HTML]{000000} \textbf{25.19}} &
  {\color[HTML]{000000} \textbf{\underline{14.38}}} &
  {\color[HTML]{000000} 23.11} &
  {\color[HTML]{000000} 32.19} \\\midrule

\end{tabular}
\caption{Performance of baseline models on different dataset settings of UIT-ViCoV19QA}
\label{tbl_overall_results}
\end{table*}

To assess the dataset's quality and illustrate the effect of having multiple paraphrased answer, we conduct and run individual experiment with four different dataset settings: one answer, two answers, three answers and finally, four paraphrased answers per question.

For Transformer, Adam optimizer is implemented with parameters as follows: $\beta_1=0.9$, $\beta_2=0.98$, $\epsilon=10^{-9}$ and $warmup\_step=2000$. This setup varies the learning rate of the model during training progress by increasing it linearly for the first 2000 training steps and decreasing it after that proportionally to the inverse square root of the step number. For other models, standard Adam optimizer with a fixed learning rate of 0.001 is applied in the training.

The training progress is configured with a batch size of 8 in 30 epochs on an NVIDIA Tesla P100 GPU via the Kaggle platform\footnote{\url{https://www.kaggle.com/}}.
After each epoch, the performance loss on the train and development sets is calculated using the Cross-Entropy Loss function. The maximum length of the model-generated output is limited to 500 tokens to reduce the generating time of repetitive loops caused by text degeneration.

\section{Experimental Results}

Table \ref{tbl_overall_results} presents the performance of our baseline models on different settings of the UIT-ViCoV19QA dataset. The final score of each metric in the table is achieved by calculating the average scores of all model-generated answers.

Evaluated using BLEU-1 and BLEU-4, the performance of models tends to improve when applying more paraphrased answers in the experiments,
though this trend behaves differently among models.
RNN-1 using two answers per question achieved the
best BLEU-1 of 26.62\% while on BLEU-4, Transformer using four answers outperforms other models
with a score of 14.38\%. Apart from RNN-1 achieves
best score when applying two answers, other models
have highest performance when training and evaluating using four answers.


In contrast with the BLEU score, the performance of models evaluated by METEOR and ROUGE-L varies significantly under different dataset settings. As shown in the table, RNN-2 and Convolutional perform well when using only one response, with scores of 33.95\% and 32.29\% respectively, while RNN-1 and Transformer need to apply more answers to achieve better scores.

Three examples of models generated responses are shown in Table \ref{tbl_error_analysis} to illustrate the generation performance comparing with the original answer.

\begin{table*}[!ht]
\renewcommand{\arraystretch}{1.2}
\footnotesize
\begin{center}
\begin{tabular}{p{0.95\linewidth}}
\hline 
\textbf{\underline{Question:}} tôi bị dị ứng với thuốc giảm đau giãn cơ . tôi uống thuốc này lúc bị sút lưng ( phồng đĩa đệm ) . vậy xin hỏi tôi có tiêm vaccine covid-19 được không ạ ? [\textbf{English:  }I am allergic to muscle relaxants. I took this medicine when I had a low back pain (bulging disc). So can I get the covid-19 vaccine?]\\

\textbf{\underline{Original Answer:}} chào anh / chị , nếu tình trạng dị ứng của anh / chị ở mức độ thông thường , không có phản vệ từ độ 2 trở lên , anh / chị vẫn có thể tiến hành tiêm chủng tại bệnh viện được . chúc anh / chị nhiều sức khỏe . trân trọng ! [\textbf{English: }hello sir / madam , if your allergy is at normal level , not having 2nd degree anaphylaxis or higher , you can still carry out vaccination at the hospital . Wish you a lot of health. Best regards !]\\

\textbf{\underline{Generated Answer:}} chào anh , theo hướng dẫn của bộ y tế , anh thuộc nhóm đối tượng cần thận trọng khi tiêm chủng vaccine covid-19 . tuy nhiên , anh cần cung cấp đầy đủ thông tin về tình trạng dị ứng của mình cho bác sĩ khám sàng lọc để được tư vấn cụ thể hơn . cảm ơn câu hỏi của anh . trân trọng ! [\textbf{English: } hello sir, according to the guidance of the Ministry of Health, you are in the group of subjects that need to be cautious when vaccinating against the covid-19 vaccine. however, you need to provide enough information about your allergy to the screening doctor for more specific advice. Thanks for your question. best regards !]\\\hline

\textbf{\underline{Question:}} ai không nên tiêm vaccine covid-19 ? [\textbf{English: }who should not get the covid-19 vaccine?]\\

\textbf{\underline{Original Answer:}} theo tổ chức y tế thế giới ( who ) , mỗi quốc gia cần chủng ngừa cho ít nhất 70 \% dân số mới đạt được miễn dịch cộng đồng . giống với các loại dược phẩm khác , vaccine covid-19 có thể gây một số tác dụng phụ và không phù hợp ở từng đối tượng nhất định . tất cả những người có phản ứng dị ứng nghiêm trọng với bất kỳ thành phần hoạt chất hoặc tá dược nào liệt kê trong mục thành phần vắc xin đều không được tiêm vắc xin . bên cạnh đó , cần thận trọng tiêm chủng với các đối tượng , như : nếu bạn đã từng có phản ứng dị ứng nghiêm trọng ( phản ứng phản vệ ) sau khi tiêm vắc xin nào đó trước đây . nếu hệ miễn dịch của bạn bị suy yếu ( suy giảm miễn dịch ) hoặc bạn đang dùng thuốc làm suy yếu hệ miễn dịch ( như corticosteroid liều cao , thuốc ức chế miễn dịch hoặc thuốc ung thư ) . nếu bạn hiện đang bị nhiễm trùng nặng với thân nhiệt cao ( trên 38 °c / 100.4 °f ) . nếu bạn có vấn đề về xuất huyết / chảy máu hoặc bầm tím , hoặc nếu bạn đang dùng thuốc làm loãng máu ( thuốc chống đông máu ) . trong trường hợp nếu bạn không chắc chắn bất kỳ điều gì bên trên , hãy trao đổi với bác sĩ hoặc chuyên gia chăm sóc sức khỏe của bạn trước khi bạn được tiêm vắc xin .  [\textbf{English: }according to the world health organization ( who ) , every country needs to immunize at least 70 \% of the population to achieve herd immunity . Like other medicines, the covid-19 vaccine can cause some side effects and is not suitable for certain people. All people who have had a severe allergic reaction to any of the active ingredients or excipients listed in the vaccine ingredients section should not get the vaccine. In addition , care should be taken in vaccinating subjects , such as : if you have ever had a severe allergic reaction ( anaphylaxis ) after receiving a certain vaccine in the past . if your immune system is weakened ( immunocompromised ) or you are taking medicine that weakens your immune system ( such as high-dose corticosteroids , immunosuppressants or cancer drugs ) . if you currently have a severe infection with a high body temperature ( above 38 °c / 100.4 °f ) . if you have bleeding / bleeding or bruising problems , or if you are taking blood thinners ( anticoagulants ) . In case if you are not sure of any of the above, talk to your doctor or healthcare professional before you receive the vaccine.]\\

\textbf{\underline{Generated Answer:}} người dân cần thực hiện nghiêm túc , chấp hành quy định của địa phương , thực hiện thông điệp 5k để bảo vệ bản thân và những người xung quanh .[\textbf{English: }people need to strictly comply with local regulations and implement the 5k message to protect themselves and those around them.]\\\hline

\textbf{\underline{Question:}} biến chủng Epsilon ( B. 1.427 / B. 1.429 ) là gì ? [\textbf{English: }what is Epsilon (B. 1,427/B. 1,429) variant?]\\

\textbf{\underline{Original Answer:}} biến chủng này có thể tự sắp xếp lại một phần của protein đột biến mà virus sử dụng để liên kết với các tế bào của cơ thể . vì thế , biến chủng epsilon có khả năng lây nhiễm cao hơn khoảng 20 \% và dễ dàng làm suy yếu hệ miễn dịch của con người . [\textbf{English: }this variant can rearrange a portion of the mutated protein that the virus uses to bind to the body's cells. Therefore, epsilon variants are about 20\% more contagious and easily weaken the human immune system.]\\

\textbf{\underline{Generated Answer:}} theo quy định tại điều 11 nghị định số 176 / 2013 / nđ-cp ngày 14/11/2013 của chính phủ về xử phạt hành chính trong lĩnh vực y tế , hành vi không thực hiện biện pháp cách ly y tế tại điều 11 nghị định số 176 / 2013 / nđ-cp ngày 14/11/2013 của chính phủ về xử phạt hành chính trong lĩnh vực y tế , hành vi có hành vi có hành vi này bị phạt tiền từ 5.000.000 đồng \hl{đến 10.000.000 đồng} \hl{đến 10.000.000 đồng} \hl{đến 10.000.000 đồng} ... [\textbf{English: }in accordance with Article 11 of the Government's Decree No. 176/2013/nd-cp dated November 14, 2013 on administrative sanctions in the health sector, failure to implement medical isolation measures in Article 11 of the Government's Decree No. 176/2013/nd-cp dated November 14, 2013 on administrative sanctions in the health sector, acts committed in this act are subject to a fine of from VND 5,000,000,000 \hl{to VND 10,000,000} \hl{to VND 10,000,000} \hl{to VND 10,000,000} ...]\\\hline
\end{tabular}
\end{center}
\caption{Examples of answers generated by the proposed models compared with the original answers\vspace{0.7cm}}
\label{tbl_error_analysis} 
\end{table*}




\subsection{Error Analysis}

From the experiment results, we determine that the Transformer trained on four answers gives the best performance among others based on BLEU-4 according to the approach of \newcite{kacupaj2021paraqa}. Twenty samples are randomly chosen from the generated answers of the model to perform error analysis. We calculate the average length and vocabulary size and count the number of POS tags in the references and generated answers. The statistics of the analysis process are shown in Table \ref{tbl_stat_analysis}.

\begin{table}[H]
    \renewcommand{\arraystretch}{1.2}
    \small
    \begin{tabular}{p{2.3cm}p{2.7cm}p{1.6cm}}\hline
     & \textbf{Original answers}  &  \textbf{Generated answers} \\\hline
     Avg. length  & \multicolumn{1}{r}{115.53}  & \multicolumn{1}{r}{75.25} \\
     Vocabulary size  & \multicolumn{1}{r}{776}  & \multicolumn{1}{r}{114}  \\
     \# POS tag  & \multicolumn{1}{r}{16}  & \multicolumn{1}{r}{12}\\\hline
    \end{tabular}
    \caption{The average length, vocabulary size and number of POS tag in 20 generated samples and their references}
    \label{tbl_stat_analysis} 
\end{table}

The vocabulary size of the output answers is almost seven times less than that of the original answers, while the average length and number of POS tags do not vary significantly. This implies that Transformer does not generalize well, and the tokens overlapping or text degeneration may have occurred in the outputs of the model.

By checking the generated results of Transformer, we notice some output answers contain overlapping phrases and do not match the question and reference answers. Further review the output of other models, various mistakes and errors have also been pointed out in the generated responses. In summary, these errors can be divided into three main groups.

\begin{itemize}
    \item The first group includes meaningless and illogical answers. In some cases, the generated answers are just a set of tokens arranged in a chaotic order.
    
    \item The second group contains meaningful and coherent answers but do not satisfy the question's requirements, as illustrated by the second example in Table \ref{tbl_error_analysis}. This type of error can be caused by the diverse Vietnamese vocabulary and grammar or long passage, which makes the model difficult to understand the context of question.
    
    \item The third group consists of text degeneration. Typically, this is a common phenomenon in sequence-to-sequence learning tasks when a word or a phrase is infinitely repeated in the generated sequence indicating that the models may not generalize well. As illustrated by the third example in Table \ref{tbl_error_analysis}, the generated response does not satisfy the question and the phrase "tới 10.000.000 đồng" (to VND 10,000,000) keeps repeated until the end of the answer.
\end{itemize}

\section{Conclusion and Future Works}
In this paper, we presented UIT-ViCoV19QA, the first community-based question answering dataset about COVID-19 for Vietnamese. Our dataset comprises 4,500 question-answer pairs with multiple paraphrased answers. The dataset's quality was evaluated through various baseline deep learning models and commonly used metrics such as BLEU, METEOR, and ROUGE-L. 

We illustrated the effect of having multiple paraphrased answers for experiments with baseline models and provided benchmark results for further research. 
RNN with Bahdanau attention achieves the best BLEU-1 and METEOR scores of 26.62\% and 25.98\% when applying two and three answers respectively. 
Transformers using four answers outperforms others on BLEU-4 with score of 14.38\%. On ROUGE-L, RNN with Luong Attention using one answer has the best performance of 33.95\%. The advantage of having multiple paraphrased answers is greatly illustrated by BLEU scores, on which three out of four models achieve the best performance when applying all four paraphrased answers. On the contrary, our experiments showed that METEOR and ROUGE-L scores do not give a clear reflection of the improvement in models performance when increase number of answer used.

Through error analysis, we showed that the performance of these models is not quite good since the generated answers contain various errors. There are several reasons for this: the diversity of the Vietnamese language, lacking a specific evaluation metric for the Vietnamese language, the long-sequence content, and the size limitation of our dataset. The dataset offers a valuable contribution to the community, providing the foundation for many research lines in the single-turn QA domain and other areas. 

In the future, UIT-ViCoV19QA can be expanded in size by collecting more relevant question-answer pairs and creating more paraphrased answers. The embedding layer of the proposed baselines can also be investigated to be replaced with pre-trained word embeddings for Vietnamese, such as PhoBERT \cite{phobert}, to improve model performance.


\subsubsection*{Acknowledgements}
This research was supported by The VNUHCM-University of Information Technology's Scientific Research Support Fund.

\bibliography{references}
\bibliographystyle{acl}

\end{document}